\documentclass[10pt,twocolumn,letterpaper]{article}

\usepackage[pagenumbers]{iccv} % arXiv

\usepackage{amsfonts}
\usepackage{amsmath}
\usepackage{amssymb}
\usepackage{bm}
\usepackage{booktabs}
\usepackage{color}
\usepackage{letltxmacro}
\usepackage{tabularray}

\UseTblrLibrary{booktabs}

\definecolor{iccvblue}{rgb}{0.21,0.49,0.74}
\usepackage[pagebackref,breaklinks,colorlinks,allcolors=iccvblue]{hyperref}

\def\paperID{10805}
\def\confName{ICCV}
\def\confYear{2025}

\gdef\papertitle{Hierarchical Modeling Approach to Fast and Accurate Table Recognition}

\title{\papertitle}

\author{Takaya Kawakatsu\\
Preferred Networks, Inc.\\
1-6-1 Otemachi, Chiyoda, Tokyo, Japan.\\
{\tt\small kat.nii.ac.jp@gmail.com}
}

\LetLtxMacro\oldeqref\eqref
\renewcommand{\eqref}[1]{Eq.~\oldeqref{eq:#1}}

\NewDocumentCommand\figref{m}{Fig.~\ref{fig:#1}}
\NewDocumentCommand\tabref{m}{Table~\ref{table:#1}}

\NewDocumentCommand\LtoR{m}{\overrightarrow{#1}}
\NewDocumentCommand\RtoL{m}{\overleftarrow{#1}}

\NewDocumentCommand\CM{}{\checkmark}
\NewDocumentCommand\trans{m}{#1^\top}
\NewDocumentCommand\softmax{m}{\mathrm{softmax}\left(#1\right)}
\NewDocumentCommand\TED{}{\mathrm{TED}}

\NewDocumentCommand\TD{}{\texttt{<td>}}
\NewDocumentCommand\TDL{}{\texttt{<td}}
\NewDocumentCommand\TDR{}{\texttt{>}}
\NewDocumentCommand\TDE{}{\texttt{</td>}}

\begin{document}
\maketitle

\begin{abstract}
The extraction and use of diverse knowledge from numerous documents is a pressing challenge in intelligent information retrieval.
Documents contain elements that require different recognition methods.
Table recognition typically consists of three subtasks, namely table structure, cell position and cell content recognition.
Recent models have achieved excellent recognition with a combination of multi-task learning, local attention, and mutual learning.
However, their effectiveness has not been fully explained, and they require a long period of time for inference.
This paper presents a novel multi-task model that utilizes non-causal attention to capture the entire table structure, and a parallel inference algorithm for faster cell content inference.
The superiority is demonstrated both visually and statistically on two large public datasets.
\end{abstract}

\section{Introduction}

Document recognition is an urgent issue because it provides full access to the vast amount of human knowledge from the past, present and future.
Large language models are capable of character recognition, and document recognition services are available.
However, their performance is still inadequate for processing diverse documents such as mathematical and chemical formulas, circuits, blueprints, celestial charts, and coats of arms.
Hidden intentions in patterns and layouts are also important for human-machine interaction.

In this paper, we focus on table image recognition, where a simple table has visible cell borders and each cell contains text, while a complex table has merged cells and/or invisible borders.
Many researchers have worked on converting such tables into machine-readable codes such as HTML~\cite{Zhong20,Li20} and LaTeX~\cite{Deng19,LaTeX21}.
Since we can utilize an optical character recognition (OCR) system to obtain cell contents, the main task is structure recognition.
Recent studies~\cite{Nassar22,VCG21,Nam23LA,Nam23GA} utilized cross attention between image features and HTML embeddings to predict HTML nodes sequentially.
This was performed in one direction, namely from top left to bottom right, making it difficult to focus on the overall structure of the table.
Bidirectional learning~\cite{Kat24} reduced this problem.

The latest studies~\cite{Nam23GA,Nam23LA,Kat24} improved the performance by using multiple decoders to learn both table structure and cell content recognition tasks.
The models~\cite{Nam23LA} successfully handled complicated tables containing hundreds of cells by using local attention~\cite{Tr20}.
The multi-cell decoder~\cite{Kat24} further improved the overall performance, which infers the contents of subsequent cells in sequence while enabling attention to the previous cells.
Actually, the multi-cell decoder was not practical because it required much inference time to predict the concatenated contents.
Furthermore, the mechanism by which the multi-task approach improved the table structure recognition task has not yet been fully explained.
For these reasons, the multi-task modeling approach has yet to spread throughout the field of document analysis.

This study extends the superior recognition performance of the multi-task model and greatly accelerates the inference speed.
The revised multi-cell decoder uses a novel inference algorithm that decodes all cell contents in parallel, speeding up inference by a factor of 10 or more.
This algorithm infers concatenated cell contents as a solution to the content length imbalance between cells.
The complete contents of the cells are obtained by repeatedly inserting the predicted next token of each cell at the end of that cell.
The combination of local attention and parallel prediction is incompatible with causal self-attention and requires the removal of attention between different cells.
This can spoil the advantage of the multi-cell decoder, and a non-causal module, a refiner, is introduced as a solution.
It focuses on the relationships between cells and organizes the structure within each cell without the need for computationally intensive autoregression.
The two methods successfully improve table recognition performance.

This paper has four major contributions.
(1) We propose a novel parallel algorithm to accelerate the table recognition model.
(2) We introduce a refiner mechanism to capture the relationship between cells and the internal structure in each cell for better performance.
(3) We demonstrate the superior performance of the model on large datasets.
(4) We provide an explanation of how the multi-task models outperform the single-task models.

\section{Related Work}

Table recognition has three subtasks, namely table structure recognition, cell bounding box recognition, and cell content recognition.
The first task outputs tokens representing rows and cells.
The second task predicts a bounding box for each cell.
The third task infers cell contents.
The structure tokens and cell content tokens are eventually combined to produce the representation in HTML~\cite{ICDAR21} or LaTeX~\cite{LaTeX21}.

It is possible to output HTML directly without executing the subtasks, but it makes sense to combine tasks at different levels of abstraction.
Since existing OCR can recognize the contents of a cell once its position is identified, and previous studies~\cite{Sch17,Pra20,Raja20,Qiao21,VCG21,Nassar22} have mainly focused on structure recognition.
They can be classified into two groups, namely bottom-up approaches and top-down approaches.

For bottom-up approaches, Schreiber et al.~\cite{Sch17} proposed table detection using Faster R-CNN~\cite{RCNN15} and table structure recognition based on semantic segmentation utilizing a fully convolutional network~\cite{FCN15}.
Raja et al.~\cite{Raja20} combined Mask R-CNN~\cite{RCNN17} and LSTM~\cite{LSTM97} to collect cell positions and then estimate the relationship between cells.
Qiao et al.~\cite{Qiao21} won his first place in the ICDAR competition~\cite{ICDAR21} by combining text, cell, row, and column recognition tasks.

For top-down approaches, a simple image caption model can be utilized, because the order of table tokens is uniquely determined, except for span tokens.
Deng et al.~\cite{Deng19} proposed a simple end-to-end model which combines a convolutional neural network (CNN) and LSTM to directly output LaTeX tokens.
Ye et al.~\cite{VCG21} developed a Transformer model using ResNet~\cite{Res15} with global context attention (GCA)~\cite{GCA19} modules for image input and two decoders for HTML tokens and cell bounding boxes.
Zhong et al.~\cite{Zhong20} used two LSTM decoders to recognize both table structure and cell contents, which is an early example of multi-task table recognition.

Remarkable progress has been made in the improvement of a complete multi-task model that learns the three subtasks of table recognition.
Ly and Takasu~\cite{Nam23GA} implemented three global attention decoders that share several attention blocks and successfully achieved accuracy comparable to the latest single-task methods without using external OCR.
They also proposed weakly supervised learning~\cite{Nam23WS} to reduce the cost of annotating tables with cell positions, and then introduced local attention~\cite{Nam23LA} to effectively recognize long tables with hundreds of cells.
Kawakatsu~\cite{Kat24} developed a bidirectional mutual learning method to enable the decoder to capture the overall structure of a table when decoding the table from top left to bottom right, and also developed a multi-cell decoder which enables cross attention between cells while decoding concatenated cell contents.

There are some public datasets of HTML annotated table images.
FinTabNet~\cite{Zheng21} contains HTML tags, cell bounding boxes, and cell contents for 112k tables.
ICDAR~\cite{ICDAR21} hosted a competition on document analysis with two tasks, namely layout recognition and table recognition.
The latter required conversion of the table images from PubTabNet~\cite{Zhong20} and the evaluation dataset for the competition into HTML.
The task accepted 30 entries, with most of the top 10 solutions using external OCR, ensemble models, and additional annotations to detect lines in each cell.
The tree edit distance on HTML excluding or including cell contents was used to score them.

Recently, scene table recognition including detection has also been investigated.
TabRecSet~\cite{Yang23} contains rotated and distorted tables for three tasks, namely table detection, table structure recognition and cell content recognition.
This also includes handwritten ones, and recognition of such tables is beyond the scope of this paper.

\section{Proposal}

\begin{figure*}[tb]
\centering
\includegraphics[width=\textwidth]{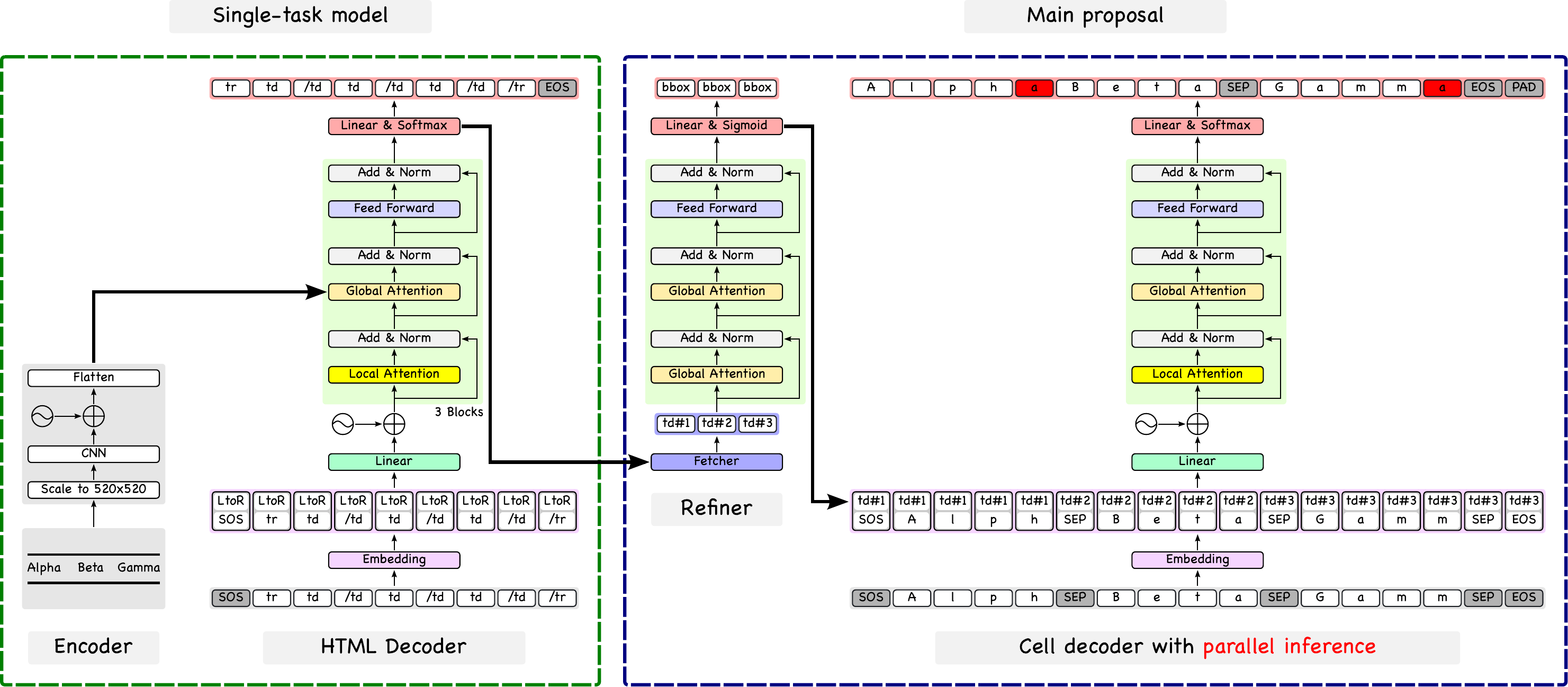}
\caption{Proposed network architecture with a refiner and parallel inference algorithm to improve cell content recognition.\label{fig:arch}}
\end{figure*}

The proposal is based on MuTabNet~\cite{Kat24}, which consists of a ResNet~\cite{Res15} encoder with GCA~\cite{GCA19} blocks for image feature extraction, two local attention~\cite{Tr20} decoders for structure and cell content recognition, respectively, and a regression layer for cell bounding boxes.
\figref{arch} shows the architecture.

Note that the main motivation of this paper is to improve cell content recognition using dense structural features, and the proposal can be easily combined with future single-task models.

\subsection{Encoder}

The encoder accepts a table image of $520\times{}520$ pixels, and extracts two-dimensional features of $65\times{}65$ pixels through a CNN backbone.
The image features are then flattened into a sequence with 4,225 pixel elements and 512 channels after two-dimensional (2D) positional encoding~\cite{Pos21}.

CNN is helpful for recognition of tiny characters, but has a poor ability to capture global contexts by mainly focusing on the local area.
As a solution, GCA~\cite{GCA19} was proposed, and we exploited TableResNetExtra~\cite{VCG21} using 26 convolutional layers and three GCA blocks.

The decoders themselves have a low ability to recognize the position of each pixel in the image and require positional encoding, where the encoded output $\bm{p}(n)$ for the $n$-th pixel is added directly to the pixel element $\bm{x}(n)$.
For table image recognition, 2D positional encoding~\cite{Pos21} may be a definitive solution, where the output for $\bm{x}(i,j)$ is given by \eqref{2d}.
\begin{equation}
\label{eq:2d}
\bm{p}_\mathrm{2D}(i,j) =
\begin{bmatrix}
\bm{p}(i) \\
\bm{p}(j) \\
\end{bmatrix}.
\end{equation}

\subsection{Decoders}

Transformer~\cite{Tr17} achieves excellent performance in various tasks.
It allows parallel learning of long sequential data and greatly reduces the vanishing gradient problem by removing recursion from training..

The key idea is called scaled dot product attention, where an attention layer first extracts a query $Q$, key $K$, and value $V$ from the input $X$ and $Y$ as defined in \eqref{QKV}.
\begin{equation}
\label{eq:QKV}
\begin{alignedat}{2}
Q &=& W_q & X, \\
K &=& W_k & Y, \\
V &=& W_v & Y, \\
\end{alignedat}
\end{equation}
with $W_q, W_k, W_v$ being the projection matrices.
The output $Z$ of the attention layer is defined by \eqref{attn}.
\begin{equation}
\label{eq:attn}
Z = W~\softmax{\frac{Q \trans{K}}{\sqrt{d}} + M} V,
\end{equation}
where $W$ is the output projection matrix and $d$ is the number of feature channels.
For causal local attention~\cite{Tr20}, the mask $M$ is given by \eqref{local}.
\begin{equation}
\label{eq:local}
M_{ij} =
\begin{cases}
\hfill 0 & 0\leq i-j \leq w, \\
- \infty & \mathrm{otherwise},
\end{cases}
\end{equation}
where $i,j$ are the indices of $Q$ and $K$, and $w$ is the width of the sliding window.
$Q\trans{K}$ provides the correlation between them, and then the softmax function calculates the attention weights, which are used to collect elements in $Y$ and update $X$ to $Z$.
This allows Transformer to capture the context and relationship between the two domains.
Attention within the same domain is called self attention, while that between two different domains is called cross-attention.

We employed the multi-head attention~\cite{Tr17}, in which the channels of $X$ and $Y$ are split into multiple groups, and each attention head calculates \eqref{attn} independently to obtain the final concatenated $Z$.

\subsubsection{HTML Decoder}

The HTML decoder consists of one embedding module, one linear layer, positional encoding, three attention blocks, and an output layer.
The decoder predicts HTML tokens shifted left (or right) from the input tokens, and iteratively expands the input sequence to obtain the complete HTML sequence.

The shift is specified by two additional vectors, LtoR and RtoL, one of which is mixed into the token embeddings via the linear layer before the attention blocks.
Each block then performs self attention, incorporates image features into the sequence through cross attention, and updates the sequence by a feed-forward layer.
The self attention uses a local mask in \eqref{local}.
Here, some skip connections and normalizations are inserted within the block.
The output from the last block is converted into HTML tokens.

Following previous work~\cite{Kat24}, we assigned some special tokens to frequently used HTML patterns.
The decoder also accepts some special control tokens.
SOS triggers inference and is inserted at the beginning of the sequence, while EOS is inserted at the end to stop inference.
PAD is inserted after EOS to equalize the sequence lengths in the mini batch.

The TD tag is tokenized as `\TDL{}', colspan, rowspan, and `\TDR{}' if it contains span attributes, or otherwise combined into a single token with the following \TDE{}.

The decoder is trained using mutual learning~\cite{Mut18}, where two student decoders predict HTML tokens in the LtoR and RtoL directions, respectively.
They are trained to teach each other without first training a teacher model that is larger than the students, as in knowledge distillation~\cite{KD15}.

Each student reverses the output sequence from the other student and calculates the Kullback-Leibler~\cite{KL51} divergence between the sequences.
Let $\LtoR{\bm{x}}$ and $\RtoL{\bm{x}}$ be the LtoR and RtoL sequences respectively, and let $p(\bm{x})$ be the ground-truth and $q(\bm{x})$ be the predicted probabilities.
The loss function $\LtoR{\mathcal{L}}$ for the LtoR output is defined by \eqref{LtoR}.
\begin{equation}
\label{eq:LtoR}
\LtoR{\mathcal{L}} =
- p(\LtoR{\bm{x}}) \log q(\LtoR{\bm{x}})
+ q(\RtoL{\bm{x}}) \log \frac{q(\RtoL{\bm{x}})}{q(\LtoR{\bm{x}})}.
\end{equation}

We reuse the same decoder for the two students by using the directional inputs, thus reducing the decoder parameters by half. 
This bidirectional mutual learning~\cite{Kat24} is helpful to effectively recognize the entire table structure without using computationally expensive ensemble learning.

\subsubsection{HTML Refiner}

The table header lacks the contextual information to predict the next structural token compared to the table footer due to causal attention.
This is unavoidable with the autoregressive inference approach, and difficult to completely resolve with bidirectional mutual learning. 
The structural features output by the HTML decoder have a major impact on the inference of cell contents.
We thus developed an HTML refiner which allows information to be shared between cells.

The refiner consists of an HTML fetcher and an attention block.
The fetcher extracts only elements that correspond to \TD{} tags from the structural features output by the HTML decoder. 
The attention block then ensures that all cells refer to each other through non-causal global attention, which has a negligible impact on the inference time, as the refiner itself does not involve iterative inference.

\subsubsection{Cell Decoder}

The cell decoder consists of an embedding module, a linear layer, positional encoding, an attention block, and an output layer.
The contents of all cells are concatenated into a single sequence, with SEP tokens placed between cells to indicate a move to the next cell.

The decoder counts the SEP tokens to identify the cell to which each character belongs.
The decoder mixes the input embedding with the structural features corresponding to the cell via the linear layer.
This may take a long time for tables with thousands of characters, and we developed the parallel inference algorithm.

The first sequence consists of an SOS token followed by SEP tokens, where the number of SEP tokens is equal to the number of found cells.
The output token of each cell is then inserted before the following SEP token.
Once a SEP token is output, no more tokens are added to that cell.
This step is repeated until all cells output the next SEP token.

Since the position of a token changes with each inference step, positional encoding must encode the positions relative to the previous SEP token rather than the absolute positions.

Local attention also needs to be improved, for many cells are at first inside the sliding window.
Once the cell contents have been expanded, other cells are outside the window and are no longer visible.
We thus introduce additional masking so that each cell cannot focus on the contents of other cells.
Since this would result in the loss of attention to other cells, we also introduced the HTML refiner.

\section{Experiments}

We conducted experiments on two public table datasets.

\subsection{Datasets}

We utilized two datasets, namely FinTabNet (FTN)~\cite{Zheng21} and PubTabNet (PTN)~\cite{Zhong20}.
They support all three subtasks, and much previous work has been compared on them.
Statistics are shown in \tabref{sets}.

\begin{table*}[tb]
\centering
\caption{The statistics of the table image datasets.\label{table:sets}}
\begin{tblr}{lrrrrrrrrr} \toprule
& \SetCell[c=3]{c}Samples & &
& \SetCell[c=3]{c}Cells per table & &
& \SetCell[c=3]{c}Characters per cell \\ \cmidrule[lr]{2-4} \cmidrule[lr]{5-7} \cmidrule[l]{8-10}
Dataset &   Train &  Valid &   Eval &  Train &  Valid &  Eval & Train & Valid &  Eval \\ \midrule
FTN     &  91,596 & 10,635 & 10,656 &  45.78 &  40.38 & 40.02 & 16.61 & 18.15 & 15.23 \\
PTN     & 500,777 &  9,115 &  9,064 &  64.86 &  66.56 & 72.52 & 13.37 & 13.54 & 12.36 \\
PTN250  & 114,111 &  2,161 &      - & 144.45 & 146.42 &     - &  8.59 &  8.52 &     - \\ \bottomrule
\end{tblr}
\end{table*}

\subsubsection{FinTabNet}

FTN is a large dataset of table images with HTML tags, cell bounding boxes, and cell contents extracted from the annual reports of S\&P 500 companies.
The dataset contains a total of around 112,000 tables and is split into training, validation and evaluation sets.
We treated the \textit{validation} set containing 10,656 tables as the evaluation set as in \cite{Zheng21,Nassar22,Nam23WS,Nam23GA,Nam23LA,Kat24}, and also extracted four subsets with a minimum of 5, 10, 15, and 20 characters per cell.

\subsubsection{PubTabNet}

PTN is a large dataset of table images with HTML tags, cell bounding boxes, and cell contents from the PubMed central open access subset.
PTN contains a total of around 510,000 images and is split into training and validation sets.
ICDAR published the evaluation set for their competition~\cite{ICDAR21}.
Note that PTN itself does not contain the evaluation set.

PTN250~\cite{Nam23LA} is a subset of tables containing at least 250 structural tokens.
It is similar in volume to FTN, but unique in that each table contains many cells, making it suitable for the ablation study.
Note that only 76 samples had more than 15 characters per cell, and their subsets were ignored.

\subsection{Metrics}

Since a typical table has a nested structure of rows, cells and cell contents, the tree edit distance-based similarity (TEDS) metric~\cite{Zhong20} is used for model evaluation.
We used an HTML parser to convert the inference results and ground truth data into trees, and calculated the TEDS using \eqref{TEDS}.
\begin{equation}
\label{eq:TEDS}
\mathrm{TEDS}(T_1, T_2) = 1 - \frac{\TED{}(T_1, T_2)}{\max(|T_1|, |T_2|)},
\end{equation}
where $T_1$ and $T_2$ are the trees, $\TED$ is the tree edit distance function that reflects insertions, deletions, and renames, and $|T|$ is the number of nodes in $T$.

To evaluate the table recognition results in detail, we use two types of TEDS metrics, structural and total.
The former is defined for the tree without cell contents, the latter for the entire tree including cell contents.

Similar to previous work~\cite{Zhong20}, we classify tables into two subsets, namely simple and complex tables.
The former are tables with no vertically or horizontally merged cells, while the latter are all other tables with merged cells.

\subsection{Implementation}

We used MMOCR~\cite{MMOCR} and modified the implementation of MuTabNet~\cite{Kat24}.
We trained it on 4 GPUs with a total batch size of 8 using the Ranger~\cite{RANGER19} optimizer.
The learning rate was 0.001, 0.0001, and 0.00001 for the first 25 epochs, next 3 epochs, and last 2 epochs, respectively.
We did not utilize early stopping and ensemble learning.

Each table image was normalized in RGB space, padded with black to preserve aspect ratio, and resized to $520\times{}520$ pixels.
The true bounding box of each cell was transformed to have a center position, width, and height with a minimum of 0 and a maximum of 1.
To ensure a fair comparison with previous work, no data augmentation was utilized.

Each token was converted into a 512-channel embedding representation.
The self-attention and cross-attention layers in the two decoders have the same 8-head architecture.
The window size for local attention was set to 300, and the token lengths for the HTML and cell decoders were limited to 800 and 8000 respectively.
We used greedy search for inference.

\subsection{Performance}

The performance of the proposal was compared to previous performance claims that clarified the method of reading cell contents, which is the main focus of this study.

\subsubsection{FinTabNet}

\begin{table}[tb]
\centering
\caption{Comparison on FinTabNet evaluation set.\label{table:fin}}
\begin{tblr}{colspec={lccc},cell{12}{3}={font=\bf},cell{6}{4}={font=\bf}} \toprule
& & \SetCell[c=2]{c}TEDS (\%) \\ \cmidrule{3-4}
Model         &                 & Structure & Total \\ \midrule
GTE           & \cite{Zheng21}  & 87.14     & -     \\
GTE (FT)      & \cite{Zheng21}  & 91.02     & -     \\
TableFormer   & \cite{Nassar22} & 96.80     & -     \\
VAST          & \cite{Huang23}  & 98.63     & 98.21 \\ \midrule
Ly et al.     & \cite{Nam23WS}  & 98.72     & 95.32 \\
Ly and Takasu & \cite{Nam23GA}  & 98.79     & -     \\
Ly and Takasu & \cite{Nam23LA}  & 98.85     & 95.74 \\
MuTabNet      & \cite{Kat24}    & 98.87     & 97.69 \\ \midrule
Proposal      &                 & 98.87     & 97.93 \\
Proposal (FT) &                 & 99.07     & 98.08 \\ \bottomrule
\end{tblr}
\end{table}

\tabref{fin} compares the TEDS scores on the evaluation set for the proposed and conventional models.
Note that the scores for the proposal are the average number of results from three independent training runs.
FT means fine tuning from PTN.

The total TEDS score of the proposal was lower than that of VAST~\cite{Huang23}, but our model outperformed all previous work in terms of structural TEDS.
VAST was unable to recognize cell contents on its own and used an external OCR~\cite{Lu21} with almost the same architecture as our proposal.
The proposed model has the potential to achieve performance comparable to VAST by increasing the number of attention blocks in the cell decoder to the same number as in the OCR~\cite{Lu21}.

Despite nearly identical architecture, the proposed model outperformed the other multi-task models~\cite{Nam23WS,Nam23GA,Nam23LA,Kat24}.
This may be due to improved cell content recognition rather than structure recognition, for the total score increased more than the structural score.
The tables in FTN tend to contain many more characters and line breaks in each cell, and the refiner can help capture the structure within each cell, as discussed in the ablation study and case study.

\subsubsection{PubTabNet}

\begin{table}[tb]
\centering
\caption{Comparison on PubTabNet validation set.\label{table:val}}
\begin{tblr}{colspec={lcccc},cell{13}{3,4,5}={font=\bf}} \toprule
& & \SetCell[c=3]{c}Total TEDS (\%) \\ \cmidrule{3-5}
Model         &                 & Simple & Complex & Total \\ \midrule
EDD           & \cite{Zhong20}  & 91.20  & 85.40   & 88.30 \\
TableFormer   & \cite{Nassar22} & 95.40  & 90.10   & 93.60 \\
SEM           & \cite{Zhang22}  & 94.80  & 92.50   & 93.70 \\
LGPMA         & \cite{Qiao21}   &     -  &     -   & 94.60 \\
VCGroup       & \cite{VCG21}    &     -  &     -   & 96.26 \\
VAST          & \cite{Huang23}  &     -  &     -   & 96.31 \\ \midrule
Ly et al.     & \cite{Nam23WS}  & 97.89  & 95.02   & 96.48 \\
Ly and Takasu & \cite{Nam23GA}  & 97.92  & 95.36   & 96.67 \\
Ly and Takasu & \cite{Nam23LA}  & 98.07  & 95.42   & 96.77 \\
MuTabNet      & \cite{Kat24}    & 98.16  & 95.53   & 96.87 \\ \midrule
Proposal      &                 & 98.34  & 95.68   & 97.04 \\ \bottomrule
\end{tblr}
\end{table}

\begin{table}[tb]
\caption{Comparison on ICDAR evaluation set.\label{table:test}}
\begin{tblr}{colspec={lcccc},cell{10}{3,4,5}={font=\bf}} \toprule
& & \SetCell[c=3]{c}Total TEDS (\%) \\ \cmidrule{3-5}
Model         &                & Simple & Complex & Total \\ \midrule
XM            & \cite{Zhang22} & 97.60  & 94.89   & 96.27 \\
VCGroup       & \cite{VCG21}   & 97.90  & 94.68   & 96.32 \\
Davar-Lab-OCR & \cite{ICDAR21} & 97.88  & 94.78   & 96.36 \\ \midrule
Ly et al.     & \cite{Nam23WS} & 97.51  & 94.37   & 95.97 \\
Ly and Takasu & \cite{Nam23GA} & 97.60  & 94.68   & 96.17 \\
Ly and Takasu & \cite{Nam23LA} & 97.77  & 94.58   & 96.21 \\
MuTabNet      & \cite{Kat24}   & 98.01  & 94.98   & 96.53 \\ \midrule
Proposal      &                & 98.10  & 95.08   & 96.62 \\ \bottomrule
\end{tblr}
\end{table}

\tabref{val} compares the TEDS scores on the validation set for the proposed and conventional models.

The proposed model outperformed all previous work for both simple and complex samples.
The proposal performed worse than VAST~\cite{Huang23} with external OCR on FTN, but much better on PTN.
This can be explained by the smaller number of characters in each table of PTN, which results in a larger penalty for incorrect structure recognition.

Despite nearly identical architecture, the proposed model outperformed the multi-task models~\cite{Nam23WS,Nam23GA,Nam23LA,Kat24}, especially for complex tables with merged cells.
The refiner may have the potential to organize the relationship between a cell and its surrounding cells, and to capture the internal structure in each cell, such as the number of characters and lines.

\tabref{test} compares the scores of the proposal with the top three solutions of the ICDAR competition~\cite{ICDAR21}, and the four multi-task models~\cite{Nam23WS,Nam23GA,Nam23LA,Kat24}.
The superior scores for both datasets indicate the high generalization performance.

\subsection{Inference Time}

\begin{table}[tb]
\centering
\caption{Inference time using an NVIDIA V100 GPU.\label{table:time}}
\begin{tblr}{colspec={cccccc}} \toprule
& & \SetCell[c=4]{c}Average inference time (s) \\ \cmidrule{3-6}
Dataset             & Parallel & HTML  & +Bbox & +Cell & OCR \\ \midrule
\SetCell[r=2]{c}FTN & -        & 0.674 & 0.679 & 3.707 & \SetCell[r=2]{c}0.081 \\
                    & \CM      & 0.787 & 0.792 & 1.300 & \\ \midrule
\SetCell[r=2]{c}PTN & -        & 1.049 & 1.056 & 4.634 & \SetCell[r=2]{c}0.102 \\
                    & \CM      & 1.150 & 1.156 & 1.506 & \\ \bottomrule
\end{tblr}
\end{table}

The proposal outperformed the multi-task approaches while being faster, and the single-task models lost their advantage in recognition.
The multi-task models~\cite{Nam23WS,Nam23GA,Nam23LA,Kat24} are based on the same model~\cite{VCG21}.
The inference time of the proposed model, the latest derivation in this series, was evaluated with parallel inference enabled or disabled.

\tabref{time} shows the inference times for FTN and PTN.
The three times in each row are the time from image acquisition to output of the table structure, refined bounding boxes, and cell contents, respectively.
The fourth is the processing time of pure OCR using SVTR~\cite{SVTR22} with distillation.

The parallel algorithm accelerated cell content inference by a factor of 10 and overall inference by a factor of 3.
The refiner performed non-autoregressive inference and was fast enough to be ignored.
It had only one block, but the number can be increased at no cost.
The autoregressive cell decoder was slower than pure OCR.
Note that the direct comparison between table structure recognition and OCR is not useful.

SVTR~\cite{SVTR22} splits the image into embedded patches, mixes and merges the patches, and derives characters directly from them without autoregression.
SVTR requires an image with exactly one line of text and must be used in conjunction with an external text detector.
Table structure must be recognized with a different model.

However, we believe that the table recognition model can be accelerated as fast as pure OCR.
The comparison results show possible table recognition speeds to be achieved in our future work.

\subsection{Ablation Study}

\begin{table*}[tb]
\centering
\caption{Ablation study of cell decoder using tables of at least 0, 5, 10, 15, and 20 characters per cell.\label{table:abl}}
\begin{tblr}{colspec={ccccccccc},cell{7}{3,4,5,6,7,8,9,10}={font=\bf}} \toprule
& & \SetCell[c=8]{c}Total TEDS (\%) \\ \cmidrule{3-10}
& & \SetCell[c=5]{c}FTN & & &
& & \SetCell[c=3]{c}PTN250 \\ \cmidrule[r]{3-7} \cmidrule[l]{8-10}
Model   & Refiner & 0+    & 5+    & 10+   & 15+   & 20+   & 0+    & 5+    & 10+   \\ \midrule
Bbox    & -       & 97.15 & 97.15 & 96.30 & 93.35 & 90.85 & 95.57 & 95.42 & 94.72 \\
Bbox    & \CM     & 97.19 & 97.19 & 96.42 & 93.58 & 91.17 & 95.72 & 95.55 & 94.68 \\ \midrule
Full    & -       & 97.87 & 97.87 & 97.44 & 95.56 & 94.08 & 95.79 & 95.61 & 95.07 \\
Full    & \CM     & 97.93 & 97.94 & 97.57 & 95.91 & 94.59 & 95.94 & 95.79 & 95.37 \\ \midrule
\end{tblr}
\end{table*}

\begin{table*}[tb]
\centering
\caption{Table recognition results for the case study with the difference shown in bold red text.\label{table:case}}
\subfloat[Accepting only bounding boxes.]{
\begin{tblr}{hlines,vlines,width=.98\textwidth,rows={font=\small},colspec={X[l]X[l]X[l]}}
2017 &
2016 &
2015 \\
Samsung Electronics Co., Ltd. &
Micron Technology, Inc. &
Intel Corporation \\
Taiwan Semiconductor Manufacturing &
Taiwan Semiconductor Manufacturing &
Samsung Electronics Co., Ltd. \\
&
&
Taiwan Semiconductor Manufacturing Company Limited
\end{tblr}
}
\\
\medskip
\subfloat[Accepting structural features.]{
\begin{tblr}{hlines,vlines,width=.98\textwidth,rows={font=\small},colspec={X[l]X[l]X[l]}}
2017 &
2016 &
2015 \\
Samsung Electronics Co., Ltd. &
Micron Technology, Inc. &
Intel Corporation \\
Taiwan Semiconductor Manufacturing \textcolor{red}{\textbf{Company Limited}} &
Taiwan Semiconductor Manufacturing \textcolor{red}{\textbf{Company Limited}} &
Samsung Electronics Co., Ltd. \\
&
&
Taiwan Semiconductor Manufacturing Company Limited
\end{tblr}
}
\end{table*}

We conducted an ablation study to evaluate the effectiveness of structural features in cell content recognition.
The model in which the cell decoder accepts all of the features from the refiner is referred to as the full model.
The model where the decoder accepts only the bounding box of each cell is called the bbox model.
We also prepared a derivative model of the refiner with the attention block removed, called the through model.

We compared the bbox model with the full model to test the need for information other than a bounding box, such as the internal structure within each cell and its relationship to other cells.
We also evaluated the through model to confirm the effect of positional accuracy on cell content recognition.

We used FTN and PTN250 for evaluation in this ablation study.
The tables in FTN had more characters per cell, while the tables in PTN250 had more cells.
The full model, which can incorporate information about the structure within each cell in addition to the position, was predicted to be effective on FTN.
The refiner was expected to improve the bounding boxes and be effective on PTN250.
We also used the subsets of different content lengths to confirm these hypotheses.

\tabref{abl} shows the results for each dataset.
Note that each row averages three training results.
Structural features were effective for FTN, and cell location refinement was effective for PTN250.
The advantage of both methods increased with text length for both datasets.
We came to the conclusion that close communication between structure recognition and cell content recognition provides superior performance, even for tables containing long and complex text.

\subsection{Case Study}

\begin{figure*}[tb]
\centering
\subfloat[Accepting only bounding boxes.]{
\begin{minipage}{.48\textwidth}
\centering
\includegraphics[width=\textwidth]{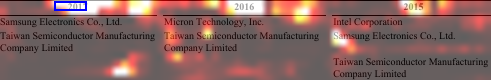} \\
\includegraphics[width=\textwidth]{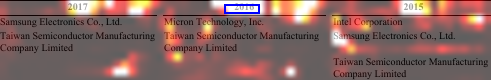} \\
\includegraphics[width=\textwidth]{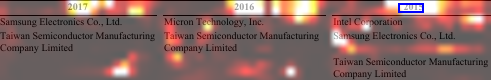} \\
\includegraphics[width=\textwidth]{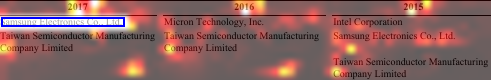} \\
\includegraphics[width=\textwidth]{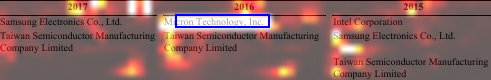} \\
\includegraphics[width=\textwidth]{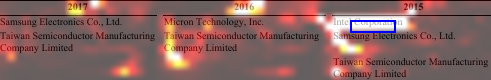} \\
\includegraphics[width=\textwidth]{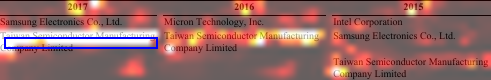} \\
\includegraphics[width=\textwidth]{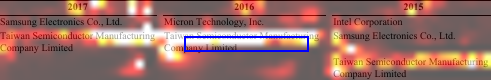} \\
\includegraphics[width=\textwidth]{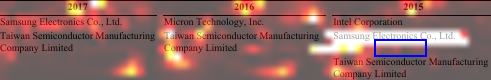} \\
\includegraphics[width=\textwidth]{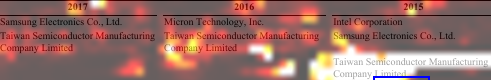}
\end{minipage}
}
\hfill
\subfloat[Accepting structural features.]{
\begin{minipage}{.48\textwidth}
\centering
\includegraphics[width=\textwidth]{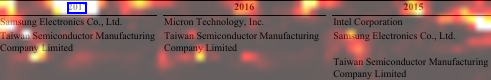} \\
\includegraphics[width=\textwidth]{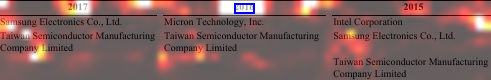} \\
\includegraphics[width=\textwidth]{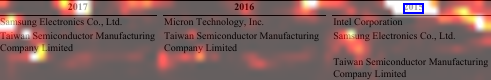} \\
\includegraphics[width=\textwidth]{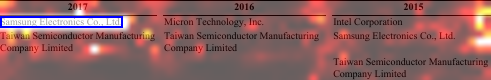} \\
\includegraphics[width=\textwidth]{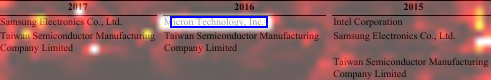} \\
\includegraphics[width=\textwidth]{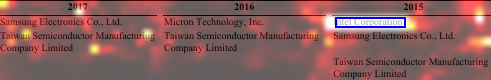} \\
\includegraphics[width=\textwidth]{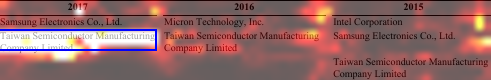} \\
\includegraphics[width=\textwidth]{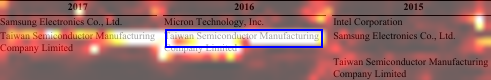} \\
\includegraphics[width=\textwidth]{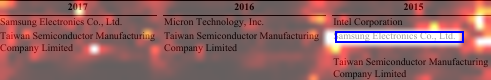} \\
\includegraphics[width=\textwidth]{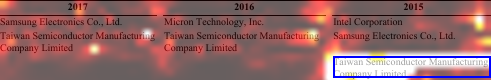}
\end{minipage}
}
\caption{Attention map from the cell decoder with the maximum shown in white and the cell bounding boxes from the refiner.\label{fig:case}}
\end{figure*}

We conducted a case study to investigate the positive impact of structural features from the HTML refiner on cell content recognition.
We focused on the cross-attention between text and image features.
We aggregated the attention weights for each cell and averaged them across all attention heads.

\figref{case} visualizes a heat map of attention weights for each decoder, with white areas attracting attention and dark areas attracting no attention.
The blue borders show the predicted bounding boxes.
\tabref{case} shows the prediction results.
Note that both models included the refiner.

The cell decoder computes the correlation between input tokens and pixels, positions the characters in the image, and predicts the next character.
Its performance should improve as attention weights become more accurate and the decoder points to the correct pixels.

For the bbox model, the cell decoder successfully located the first letter of each cell, but the cell bounding boxes were rarely correct.
On the other hand, the decoder paid attention not only to the target cells, but also to the other cells around them.
These observations indicate that the decoder adjusted the provided bounding boxes on its own, rather than relying on them completely.

The decoder often stopped at the end of the first line and left the second line unread, even though the line overlapped the bounding box.
It may be difficult for the cell decoder to distinguish between a line within a cell and a table row.

For the full model, the cell bounding boxes and attention were accurate enough to read all lines and characters within each cell.
The structural features provided the decoder with accurate information about the cell contents, and the refiner appeared to be more rigorously trained than the bbox model to provide correct information.
This shows the effectiveness of sharing dense structural features between subtasks.

\section{Conclusion}

We improved a multi-task table recognition model to realize a higher level of accuracy and a faster speed than in previous work.
The model consists of an image feature extractor, two decoders for the two tasks of table structure recognition and cell content recognition, and a refiner placed between them.

The proposed parallel inference algorithm accepts a long sequence combining the contents of all cells.
The predicted token for each cell is inserted between the cells and the cell contents are inferred in parallel.
The proposed combination of cell-wise positional encoding and local attention achieves this while reducing memory consumption.

The refiner is designed to address the problem that earlier cells are less likely to capture the context of the entire table than later cells due to unidirectional prediction.
It promotes information sharing between cells, which is disrupted by the cell-wise local attention.

We will integrate the method with the non-autoregressive OCR models in future work, which will allow for even more accurate and faster table analysis.
The main contribution of this paper is to improve cell content recognition using dense structural features, and the proposal can be easily combined with future single-task models.

We demonstrated the effectiveness of the method on two large-scale datasets.
We further analyzed the benefits of the multi-task model for the first time through the ablation study and case study.

{\small
\bibliographystyle{iccv}
\bibliography{iccv25}

@inproceedings{Kat24,
	author={Takaya Kawakatsu},
	title={Multi-Cell Decoder and Mutual Learning for Table Structure and Character Recognition},
	booktitle={Document Analysis and Recognition - ICDAR 2024},
	publisher={Springer Nature Switzerland},
	year={2024},
	pages={389--405},
}

@inproceedings{Nam23LA,
	author={Nam Tuan Ly and Atsuhiro Takasu},
	title={An End-to-End Local Attention Based Model for Table Recognition},
	booktitle={International Conference on Document Analysis and Recognition ({ICDAR})},
	year={2023},
	pages={20--36},
}

@inproceedings{Nam23GA,
	author={Nam Tuan Ly and Atsuhiro Takasu},
	title={An End-to-End Multi-Task Learning Model for Image-based Table Recognition},
	booktitle={International Joint Conference on Computer Vision, Imaging and Computer Graphics Theory and Applications ({VISIGRAPP})},
	year={2023},
	pages={626--634},
}

@inproceedings{Nam23WS,
	author={Nam Tuan Ly and Atsuhiro Takasu and Phuc Nguyen and Hideaki Takeda},
	title={Rethinking Image-Based Table Recognition Using Weakly Supervised Methods},
	booktitle={International Conference on Pattern Recognition Applications and Methods ({ICPRAM})},
	year={2023},
	pages={872--880},
}

@inproceedings{Res15,
	author={Kaiming He and Xiangyu Zhang and Shaoqing Ren and Jian Sun},
	booktitle={IEEE Conference on Computer Vision and Pattern Recognition ({CVPR})},
	title={Deep Residual Learning for Image Recognition},
	year={2016},
	pages={770--778},
}

@inproceedings{GCA19,
	author={Yue Cao and Jiarui Xu and Stephen Lin and Fangyun Wei and Han Hu},
	booktitle={{IEEE/CVF} International Conference on Computer Vision Workshop ({ICCVW})},
	title={{GCNet}: Non-Local Networks Meet Squeeze-Excitation Networks and Beyond},
	year={2019},
	pages={1971--1980},
}

@inproceedings{RCNN15,
	author={Shaoqing Ren and Kaiming He and Ross Girshick and Jian Sun},
	title={Faster {R-CNN}: Towards Real-Time Object Detection with Region Proposal Networks},
	booktitle={Advances in Neural Information Processing Systems},
	year={2015},
	volume={28},
}

@inproceedings{RCNN17,
	author={Kaiming He and Georgia Gkioxari and Piotr Doll{\'{a}}r and Ross Girshick},
	title={Mask {R-CNN}},
	booktitle={IEEE International Conference on Computer Vision (ICCV)},
	year={2017},
	pages={2980--2988},
}

@inproceedings{FCN15,
	author={Jonathan Long and Evan Shelhamer and Trevor Darrell},
	title={Fully convolutional networks for semantic segmentation},
	booktitle={{IEEE} Conference on Computer Vision and Pattern Recognition ({CVPR})},
	year={2015},
	pages={3431--3440},
}

@article{LSTM97,
	author={Sepp Hochreiter and Jürgen Schmidhuber},
	title={Long Short-term Memory},
	journal={Neural computation},
	year={1997},
	volume={9},
	pages={1735--80},
}

@inproceedings{Tr17,
	author={Ashish Vaswani and Noam Shazeer and Niki Parmar and Jakob Uszkoreit and Llion Jones and Aidan N. Gomez and Lukasz Kaiser and Illia Polosukhin},
	title={Attention is All you Need},
	booktitle={Advances in Neural Information Processing Systems},
	year={2017},
	volume={30},
	pages={6000--6010},
}

@misc{Tr20,
	title={{Longformer}: The Long-Document Transformer},
	author={Iz Beltagy and Matthew E. Peters and Arman Cohan},
	year={2020},
}

@InProceedings{Pos21,
	author={Wenqi Zhao and Liangcai Gao and Zuoyu Yan and Shuai Peng and Lin Du and Ziyin Zhang},
	title={Handwritten Mathematical Expression Recognition with Bidirectionally Trained Transformer},
	booktitle={International Conference on Document Analysis and Recognition ({ICDAR})},
	year={2021},
	pages={570--584},
}

@misc{KD15,
	author={Geoffrey Hinton and Oriol Vinyals and Jeff Dean},
	title={Distilling the Knowledge in a Neural Network},
	year={2015},
}

@inproceedings{Mut18,
	author={Ying Zhang and Tao Xiang and Timothy M. Hospedales and Huchuan Lu},
	booktitle={{IEEE/CVF} Conference on Computer Vision and Pattern Recognition ({CVPR})},
	title={Deep Mutual Learning},
	year={2018},
	pages={4320--4328},
}

@article{KL51,
	author={Solomon Kullback and Richard A. Leibler},
	title={On Information and Sufficiency},
	journal={The Annals of Mathematical Statistics},
	volume={22},
	number={1},
	year={1951}
}

@misc{RANGER19,
	author={Less Wright},
	title={Ranger - a synergistic optimizer},
	year={2019},
	url={https://github.com/lessw2020/Ranger-Deep-Learning-Optimizer}
}

@inproceedings{ICDAR21,
	author={Antonio Jimeno Yepes and Peter Zhong and Douglas Burdick},
	title={{ICDAR} 2021 Competition on Scientific Literature Parsing},
	booktitle={International Conference on Document Analysis and Recognition ({ICDAR})},
	year={2021},
	pages={605--617},
}

@inproceedings{LaTeX21,
	author={Pratik Kayal and Mrinal Anand and Harsh Desai and Mayank Singh},
	title={{ICDAR} 2021 Competition on Scientific Table Image Recognition to {LaTeX}},
	booktitle={International Conference on Document Analysis and Recognition ({ICDAR})},
	year={2021},
	pages={754--766},
}

@inproceedings{Pra20,
	author={Devashish Prasad and Ayan Gadpal and Kshitij Kapadni and Manish Visave and Kavita Sultanpure},
	title={{CascadeTabNet}: An approach for end to end table detection and structure recognition from image-based documents},
	booktitle={{IEEE/CVF} Conference on Computer Vision and Pattern Recognition Workshops ({CVPRW})},
	year={2020},
	pages={2439--2447},
}

@inproceedings{Raja20,
	author={Sachin Raja and Ajoy Mondal and C. V. Jawahar},
	title={Table Structure Recognition Using Top-Down and Bottom-Up Cues},
	booktitle={Computer Vision -- {ECCV}},
	year={2020},
	pages={70--86},
}

@inproceedings{Sch17,
	author={Sebastian Schreiber and Stefan Agne and Ivo Wolf and Andreas Dengel and Sheraz Ahmed},
	title={{DeepDeSRT}: Deep Learning for Detection and Structure Recognition of Tables in Document Images},
	booktitle={International Conference on Document Analysis and Recognition ({ICDAR})},
	year={2017},
	pages={1162--1167},
}

@inproceedings{Zhong20,
	author={Xu Zhong and Elaheh ShafieiBavani and Antonio Jimeno Yepes},
	title={Image-Based Table Recognition: Data, Model, and Evaluation},
	booktitle={Computer Vision -- {ECCV}},
	year={2020},
	pages={564--580},
}

@inproceedings{Li20,
	title={{T}able{B}ank: Table Benchmark for Image-based Table Detection and Recognition},
	author={Minghao Li and Lei Cui and Shaohan Huang and Furu Wei and Ming Zhou and Zhoujun Li},
	booktitle={Language Resources and Evaluation Conference (LREC)},
	year={2020},
	pages={1918--1925},
	ISBN={979-10-95546-34-4},
}

@inproceedings{Deng19,
	author={Yuntian Deng and David Rosenberg and Gideon Mann},
	title={Challenges in End-to-End Neural Scientific Table Recognition},
	booktitle={International Conference on Document Analysis and Recognition ({ICDAR})},
	year={2019},
	pages={894--901},
}

@inproceedings{Qiao21,
	author={Liang Qiao and Zaisheng Li and Zhanzhan Cheng and Peng Zhang and Shiliang Pu and Yi Niu and Wenqi Ren and Wenming Tan and Fei Wu},
	title={{LGPMA}: Complicated Table Structure Recognition with Local and Global Pyramid Mask Alignment},
	booktitle={International Conference on Document Analysis and Recognition ({ICDAR})},
	year={2021},
	pages={99--114},
}

@inproceedings{Nassar22,
	author={Ahmed Nassar and Nikolaos Livathinos and Maksym Lysak and Peter Staar},
	title={{TableFormer}: Table Structure Understanding with Transformers},
	booktitle={{IEEE/CVF} Conference on Computer Vision and Pattern Recognition ({CVPR})},
	year={2022},
	pages={4604--4613},
}

@misc{VCG21,
	author={Jiaquan Ye and Xianbiao Qi and Yelin He and Yihao Chen and Dengyi Gu and Peng Gao and Rong Xiao},
	title={{PingAn}-{VCGroup}'s Solution for {ICDAR} 2021 Competition on Scientific Literature Parsing Task {B}: Table Recognition to {HTML}},
	year={2021},
}

@article{Lu21,
	author={Ning Lu and Wenwen Yu and Xianbiao Qi and Yihao Chen and Ping Gong and Rong Xiao and Xiang Bai},
	title={{MASTER}: Multi-aspect non-local network for scene text recognition},
	journal={Pattern Recognition},
	year={2021},
	volume={117},
	pages={107980},
}

@article{Zhang22,
	author={Zhenrong Zhang and Jianshu Zhang and Jun Du and Fengren Wang},
	title={Split, Embed and Merge: An accurate table structure recognizer},
	journal={Pattern Recognition},
	year={2022},
	volume={126},
	pages={108565},
}

@inproceedings{Zheng21,
	author={Xinyi Zheng and Douglas Burdick and Lucian Popa and Xu Zhong and Nancy Xin Ru Wang},
	title={Global Table Extractor ({GTE}): A Framework for Joint Table Identification and Cell Structure Recognition Using Visual Context},
	booktitle={IEEE Winter Conference on Applications of Computer Vision ({WACV})},
	year={2021},
	pages={697--706},
}

@inproceedings{Huang23,
	author={Yongshuai Huang and Ning Lu and Dapeng Chen and Yibo Li and Zecheng Xie and Shenggao Zhu and Liangcai Gao and Wei Peng},
	title={Improving Table Structure Recognition with Visual-Alignment Sequential Coordinate Modeling},
	booktitle={{IEEE/CVF} Conference on Computer Vision and Pattern Recognition (CVPR)},
	year={2023},
	pages={11134-11143}
}

@inproceedings{SVTR22,
	title={{SVTR}: Scene Text Recognition with a Single Visual Model},
	author={Yongkun Du and Zhineng Chen and Caiyan Jia and Xiaoting Yin and Tianlun Zheng and Chenxia Li and Yuning Du and Yu-Gang Jiang},
	booktitle={International Joint Conference on Artificial Intelligence},
	year={2022},
	pages={884--890},
}

@article{Yang23,
	author={Fan Yang and Lei Hu and Xinwu Liu and Shuangping Huang and Zhenghui Gu},
	title={A Large-scale Dataset for End-to-end Table Recognition in the Wild},
	journal={Scientific Data},
	year={2023},
	volume={10},
	number={1},
	pages={110},
}

@misc{MMOCR,
	author={{OpenMMLab}},
	title={{MMOCR}},
	url={https://github.com/open-mmlab/mmocr},
}
}

\end{document}